\documentclass[runningheads]{llncs}

\usepackage[T1]{fontenc}
\usepackage{graphicx,verbatim}
\usepackage{booktabs}
\usepackage{amsmath}
\usepackage{amssymb}
\usepackage{multirow}
\usepackage{enumitem}
\usepackage[colorlinks,linkcolor=red,anchorcolor=blue,citecolor=blue,urlcolor=magenta]{hyperref}
\usepackage{tcolorbox}
\tcbuselibrary{skins}
\definecolor{systembg}{RGB}{240,248,255}
\definecolor{answerbg}{RGB}{255,245,238}

\begin{document}

\title{Enhancing Brain MRI Anomaly Detection and Reasoning with ROI Rethink and Synthetic Data}
\titlerunning{BrReMark}

\author{Shangkun Li\inst{1}\textsuperscript{*} \and 
        Jie Xu\inst{1}\textsuperscript{*} \and 
        Yi Guo\inst{1} \and 
        Zeju Li\inst{1}\textsuperscript{\dag} \and 
        Yuanyuan Wang\inst{1}\textsuperscript{\dag}}


\authorrunning{S. Li et al.}

\institute{College of Biomedical Engineering, Fudan University, Shanghai, China \\
\email{\{22307130255, xujie23\}@m.fudan.edu.cn, \{guoyi, zejuli, yywang\}@fudan.edu.cn}}

{
  \renewcommand{\thefootnote}{} 
  \footnotetext{$^*$ Equal contribution. $^\dag$ Corresponding authors.\\
  Code is available at \url{https://github.com/fdu-farm/BrReMark}.}
}

\maketitle

\begin{abstract}
Medical vision-language models typically generate diagnoses through single-pass inference without indicating which image regions support their conclusions. This lack of spatial grounding limits clinical utility: outputs cannot be audited, and models may hallucinate findings on normal scans. We present \textbf{BrReMark} (\textbf{Br}ain \textbf{Re}think via ROI \textbf{Mark}ing), a framework that introduces explicit region marking into brain MRI diagnosis. The model first generates hypotheses about potential abnormalities and grounds them through explicit bounding box marking, then verifies conclusions by re-examining the marked evidence. Training combines supervised fine-tuning on structured reasoning trajectories with reinforcement learning using a composite reward over localization accuracy and diagnostic reasoning. Furthermore, we integrate a domain randomization-based pathology synthesis augmentation strategy to improve the model's generalizability to out-of-distribution (OOD) data. On internal benchmark, BrReMark improves mAP$_{50}$ from 0.74\% to 37.54\% compared to the base model, while achieving 21.57 Clinical F1 and 45.26\% diagnostic accuracy. On NOVA OOD benchmark, it also achieves competitive overall performance with a 45.7$\%$ reduction in false positives compared to the state-of-the-art, indicating reduced hallucination on rare pathologies. These findings suggest that explicit hypothesis-verification grounding is a practical path toward trustworthy open-ended brain MRI diagnosis across both in-distribution and OOD settings.
\keywords{Brain MRI \and Medical VLMs \and Reasoning}
\end{abstract}

\section{Introduction}

Trustworthy AI-assisted diagnosis requires not only accurate predictions but also auditable reasoning that clinicians can verify~\cite{liu2024survey}. Medical vision-language models (VLMs) (e.g., LLaVa-Med~\cite{Llava-med}, HuatuoGPT-Vision~\cite{chen2024towards}, and Lingshu~\cite{lingshu}) have achieved strong performance on radiology dialogue and visual question answering, yet their clinical utility remains limited by a fundamental trustworthiness gap: models generate conclusions without indicating which image regions support the diagnosis, making outputs impossible to audit and prone to hallucination on normal scans~\cite{zhang2026data}. This issue is particularly acute for brain MRI, where lesion location directly informs differential diagnosis and the prevalence of rare neurological diseases introduces severe out-of-distribution challenges~\cite{bercea2026nova}.

To bridge this trustworthiness gap, the emerging "Thinking with Images" paradigm~\cite{fan2026grit,openthinkimg,deepeyes} offers a promising solution by shifting from passive observation to active perception. Unlike traditional models that rely on single-pass inference, this paradigm facilitates iterative interaction with visual data. This multi-turn reasoning process allows the model to formulate intermediate hypotheses, zoom in on diagnostically salient regions, and perform self-correction. Consequently, the final diagnostic decision is supported by a verifiable chain of visual evidence, aligning the model's behavior with clinical standards. Unlike fTSPL, which generates text from fMRI activations to assist representation learning and prediction~\cite{wang2024ftspl}, BrReMark explicitly grounds lesion locations in structural MRI and produces auditable ROI markings to support diagnosis.

Despite recent advancements, several critical limitations hinder its clinical application in brain MRI diagnosis. Primarily, prevailing benchmarks like VQA-RAD~\cite{vqa-rad} and SLAKE~\cite{slake} emphasize closed-ended accuracy, which inherently misaligns with the open-set nature of real-world clinical scenarios. Consequently, open-ended anomaly detection remains largely underexplored. Furthermore, constrained by these closed-set paradigms and limited data, existing models tend to overfit to known distributions, severely compromising their ability to generalize to out-of-distribution (OOD) data~\cite{bercea2026nova}. Developing novel methods capable of handling open-set clinical complexities is essential for real-world VLM deployment.

To this end, we present \textbf{BrReMark} (\textbf{B}rain \textbf{Re}think via ROI \textbf{Mark}ing), a novel "Think with Images" framework that tackles open-set OOD clinical challenges by explicitly mirroring the human diagnostic workflow. Specifically, BrReMark employs a progressive two-stage training paradigm: Supervised Fine-Tuning (SFT) followed by Group Relative Policy Optimization (GRPO~\cite{grpo})-based Reinforcement Learning (RL). Our main contributions are fourfold:
\begin{enumerate}
    \item \textbf{Interactive "Mark-and-Rethink" Trajectory via SFT}:  Rather than relying on standard QA pairs, this design elicits an auditable reasoning process: BrReMark first hypothesizes and explicitly marks suspicious regions of interest (ROIs), and subsequently uses these visual cues to "rethink" and verify initial findings. This explicit trajectory empowers it to navigate complex open-set clinical scenarios inherently neglected by closed-set tasks.
    \item \textbf{Clinical-Aligned RL Optimization}: To facilitate free-form, open-ended diagnosis, we design a multi-component reward function encompassing localization accuracy, semantic correctness, and clinical safety. Driven by GRPO, this composite reward effectively aligns the VLM's generative capabilities with stringent clinical requirements, elevating its performance from rigid closed-ended metrics to reliable diagnostic narratives.
    \item \textbf{Synthesis-Driven OOD Robustness}: We address the scarcity of rare clinical lesions by extending SynthSeg with pathological data augmentation, compelling the model to localize structural abnormalities during the RL stage. This strategy enhances the model’s robustness on unseen OOD cases, mirroring the complexities of real-world medical data.
    \item \textbf{Comprehensive Evaluation}: We introduce a new two-turn interactive brain MRI dataset designed to realize BrReMark framework, representing a comprehensive integration of seven open-source datasets. We thoroughly validated BrReMark against internal and external benchmarks; it achieves superior results compared to other models within the same capacity class.
\end{enumerate}

\begin{figure}[t]
\centering
\includegraphics[width=\textwidth]{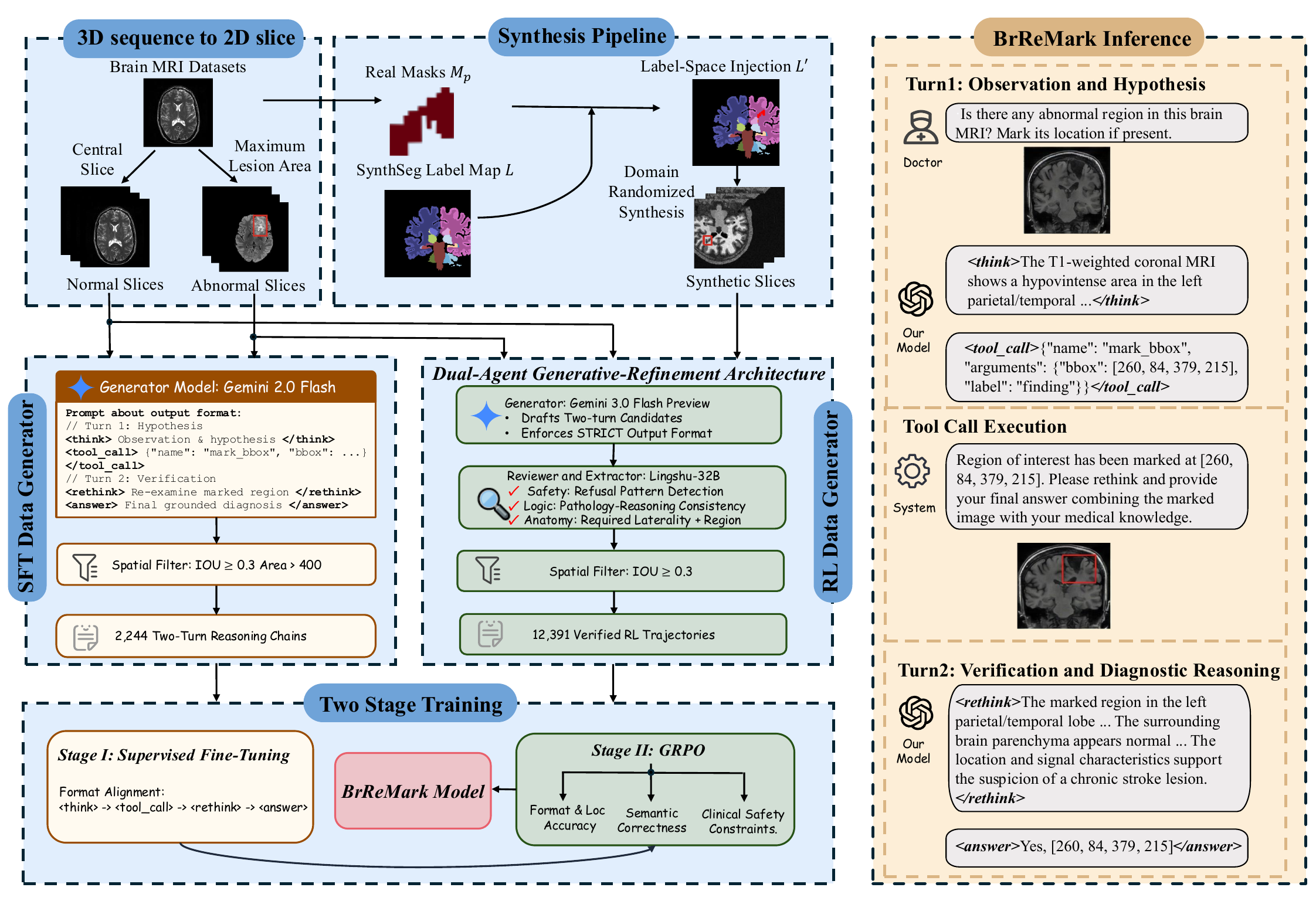}
\caption{\textbf{Left: } Data curation and training pipeline. Training data combines real abnormal slices, normal controls, and SynthSeg-generated slices. \textbf{Right: } BrReMark Inference pipeline. Given a brain MRI and query, the model first generates a hypothesis and invokes \texttt{mark\_bbox} to localize suspicious regions (Turn 1), then re-examines the marked image to verify findings and produce a grounded diagnosis (Turn 2). }
\label{fig:framework}
\end{figure}


\section{Methods}
\subsection{Overview}
We formulate brain MRI diagnosis in the BReMark framework as a two-turn visual dialogue (Fig.~\ref{fig:framework}). Given an image $\mathcal{I}$ and query $q$, the VLM $\pi_\theta$ first proposes a hypothesis $h$ and a bounding box $b$ (where $b = \texttt{null}$ for normal images): $o_1 = (h, b) \sim \pi_\theta(\cdot \mid \mathcal{I}, q)$. The box is then rendered to produce a marked image $\mathcal{I}_m$. In the second turn, the model verifies its initial findings using this augmented input to output $o_2 = (v, y) \sim \pi_\theta(\cdot \mid \mathcal{I}, q, o_1, \mathcal{I}_m)$, yielding the verification reasoning $v$ and the final diagnostic answer $y$.

During the training process, we employ a two-stage optimization strategy. Stage 1 utilizes SFT to instruct the VLM to adhere to the aforementioned two-turn diagnostic format. Subsequently, stage 2 leverages RL to ensure the model's reasoning and outputs align with human preferences.

\subsection{Open Set Data Curation}
\label{sec:open-set}
To reflect real-world clinical challenges, which are inherently open-ended rather than closed-ended (e.g., multiple-choice), we curate a comprehensive open-set dataset for both the VLM's SFT and RL stages. We initially compile approximately 3,700 cases across five modalities (T1, T1-Gd, T2, FLAIR, DWI) and three anatomical planes (axial, coronal, sagittal). These are sourced from six abnormal datasets (BraTS-GLI/MEN/PED~\cite{baid2021rsna}, UPENN-GBM~\cite{bakas2021multi}, ATLAS~\cite{liew2022large}, ISLES~\cite{hernandez2022isles}) and one healthy control dataset (IXI~\cite{ixi}). During preprocessing, for each abnormal volumetric case, we extract a \(480 \times 480\) 2D slice containing the maximum lesion area based on the ground-truth mask, and generate its minimum bounding box for precise spatial localization. For normal cases from the IXI, we extract central 2D slices and explicitly set the bounding box to \texttt{null}.

Building upon this shared data pool, we first construct the SFT dataset. 
Since the SFT stage primarily requires learning format compliance, we prompt Gemini 2.0 Flash~\cite{gemini2.0} with each image and its structured priors. This explicit guidance constrains the LLM to generate clinically grounded reasoning chains, directly yielding the two-turn interactive format incorporating \texttt{<think>}, \texttt{<tool\_call>}, and \texttt{<rethink>} tags, as illustrated in Fig.~\ref{fig:data_format}(a). 

RL stage necessitates a strictly structured QA format, shown in Fig.~\ref{fig:data_format}(b). 
To achieve this, we propose a \textbf{Dual-Agent Generative-Refinement Architecture} powered by the more advanced Gemini 3.0 Flash Preview~\cite{gemini3.0}. Within this framework, a \textit{Generator Agent} first produces an initial response encompassing the full cognitive sequence. Subsequently, a \textit{Reviewer Agent} parses this output, extracting core clinical elements to populate the required template. RL data is further validated utilizing Lingshu-32B, which filters the structured outputs based on three criteria: (1) \textit{Refusal Detection} (eliminating patterns like ``I cannot''); (2) \textit{Semantic Consistency} (ensuring logical coherence across reasoning phases); and (3) \textit{Anatomical Standardization} (mandating precise spatial descriptions). After screening, we yield 12,391 high-quality structured RL samples. 

\subsection{Stage-wise Synthetic Pathology Injection for Generalization}
\label{sec:synthetic}

To enhance OOD generalization, we introduce a synthetic pathology pipeline governed by a strict \textbf{Stage-Wise Data Routing} strategy. To avoid clinical hallucinations, synthetic data is excluded from SFT. It is used only in RL for spatial targeting, with diagnostic reasoning masked to ensure the model’s foundational policy remains grounded in real-world physiology.

Specifically, to construct synthetic abnormal MRI volumes for the RL phase and improve diagnostic generalization to OOD data, we extract real lesion masks from the 6 aforementioned anomaly datasets and integrate them with healthy brain anatomical label maps provided by SynthSeg~\cite{billot2023synthseg} (where each structure receives a distinct discrete label). Our synthesis pipeline comprises two stages:\\
\textbf{Stage 1: Label-Space Pathology Injection.} Given a healthy brain label map from SynthSeg, we inject real lesion masks into this label space. To determine valid implantation sites, we employ Euclidean Distance Transform (EDT) weighted sampling within internal brain tissues (e.g., white matter, cortex). This sampling strategy explicitly biases toward deep brain regions to prevent lesion boundaries from exceeding normal brain tissue margins. \\
\textbf{Stage 2: Domain-Randomized Synthesis.}
We utilize SynthSeg generative model to synthesize images while maintaining lesion intensity distributions based on statistical analysis of real patient data. For a specific pathology, its intensity sampling prior is bounded by robust statistics (median $\pm$ $2\times$MAD) derived from the real datasets. T1 tumor utilizes $\mu \in [88, 195]$, whereas DWI ischemic lesions employ $\mu \in [207, 277]$ to authentically reflect their characteristic hyperintensity. 


\subsection{Training Strategy}
\noindent\textbf{Stage 1: Supervised Fine Tuning.}
The primary objective is to teach the model the foundational two-turn paradigm and the structural syntax of the two-turn dialogue (e.g., \texttt{<think>}, \texttt{<tool\_call>}). 
For a given image $\mathcal{I}$, query $q$, and target sequence $o^* = (o_1, o_2)$, we optimize the model parameters $\theta$ using standard autoregressive cross-entropy loss:
\begin{equation}
 \mathcal{L}_{\text{SFT}} = - \log \pi_\theta(o_1 \mid \mathcal{I}, q) - \log \pi_\theta(o_2 \mid \mathcal{I}, q, o_1, \mathcal{I}_m). 
\end{equation}
By minimizing this loss, the model learns to first generate the hypothesis and bounding box ($o_1$), then verify its findings using the marked image $\mathcal{I}_m$ ($o_2$).

\begin{figure}[t]
\centering
\includegraphics[width=\textwidth]{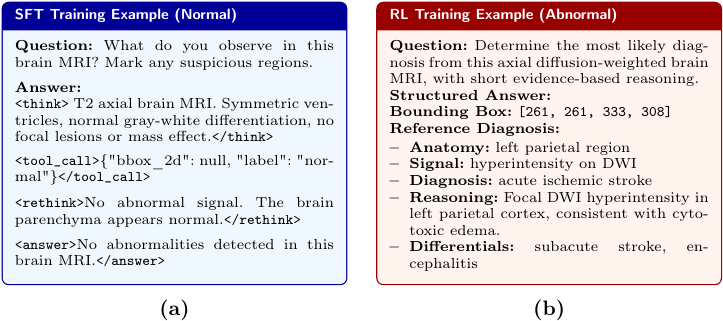}
\caption{Training data formats for BrReMark. (a) SFT example demonstrating the two-turn reasoning format with \texttt{<think>} and \texttt{<tool\_call>} tags for normal cases. (b) RL example showing structured answer format with ground-truth bounding box and reference diagnosis used for multi-component reward computation during GRPO training.}
\label{fig:data_format}
\end{figure}

\noindent\textbf{Stage 2: Reinforcement Learning.}
We design a multi-component reward function encompassing localization accuracy, semantic correctness, and clinical safety. For each input pair $(\mathcal{I}, q)$, we sample a group of $G$ candidate responses. The composite reward for the $i$-th candidate is defined as $r_i = r_{\text{fmt}} + r_{\text{loc}} + r_{\text{llm}}$:

\begin{itemize}[left=1em, itemsep=0pt, parsep=0pt]
    \item \textbf{Format \& Localization Accuracy:} To maintain structural integrity, $r_{\text{fmt}} \in [0, 0.4]$ evaluates reasoning tag completeness. For spatial precision, $r_{\text{loc}} \in [0, 0.5]$ computes bounding box accuracy via $\min(\text{IoU}, 0.5)$. 
    \item \textbf{Semantic Correctness:} To robustly evaluate diagnostic narratives, $r_{\text{llm}} \in [0, 1.0]$ utilizes Lingshu-32B~\cite{lingshu} as an LLM-as-a-judge~\cite{llm-as-a-judge}. It assesses the diagnostic validity of the generated text against the ground-truth (Fig.~\ref{fig:data_format}(b)).
\end{itemize}


\paragraph{Clinical Safety Constraints.} To mitigate clinical hallucinations and guarantee reliability, we enforce three critical safety mechanisms acting as reward gatekeepers: (1) \textit{Modality Gating:} Nullifies the semantic reward ($r_{\text{llm}} = 0$) if the MRI sequence is misidentified; (2) \textit{Hallucination Penalty:} Sets the total reward $r_i = 0$ if lesions are fabricated on healthy brain images; (3) \textit{Synthetic Masking:} $r_{\text{llm}}$ is excluded for synthetic samples to prevent artificial noise. Synthetic samples are also flagged in the prompt to distinguish them from real cases.

\paragraph{Optimization Objective.} To update the model without requiring an external value network, GRPO normalizes the rewards within each group to compute the advantage $\hat{A}_i = (r_i - \mu_r) / \sigma_r$. The VLM policy $\pi_\theta$ is then optimized by maximizing the following objective:
\begin{equation}
    \small
    \mathcal{J}_{\text{GRPO}}(\theta) = \mathbb{E} \left[ \frac{1}{G} \sum_{i=1}^G \left( \min \left( \rho_i \hat{A}_i, \text{clip}(\rho_i, 1-\epsilon, 1+\epsilon) \hat{A}_i \right) - \beta \mathbb{D}_{\text{KL}}(\pi_\theta \| \pi_{\text{ref}}) \right) \right],
\end{equation}
where $\rho_i = \frac{\pi_\theta(o_i \mid \mathcal{I}, q)}{\pi_{\text{ref}}(o_i \mid \mathcal{I}, q)}$ represents the probability ratio of the current policy $\pi_\theta$ over the reference SFT model $\pi_{\text{ref}}$, $\epsilon$ limits the update step, and $\beta$ controls the KL divergence penalty to prevent policy degradation.


\section{Experiments and Results}

\subsection{Benchmark} \label{sec:benchmark} 
We evaluate BrReMark across three clinical tasks: Anomaly Localization, Image Description, and Differential Diagnosis. Our curated dataset (Sec.~\ref{sec:open-set}) is partitioned $70\%/10\%/20\%$ for training, validation, and testing. For description and diagnosis, we use expert reports from RadGenome-Brain\_MRI~\cite{lei2025interpretable} as ground truth for the ISLES and BraTS test subsets. For OOD evaluation, we employ NOVA~\cite{bercea2026nova}, comprising 906 Eurorad scans across 281 rare pathologies. Excluded from all training phases, NOVA introduces substantial distribution shifts and long-tail anomalies to test the model's robustness to unseen diseases.

\subsection{Experimental Setup}

\noindent\textbf{Baselines and Implementation.}
We compare BrReMark against representative VLMs across three paradigms: (1) \textit{General Domain}: Qwen2.5-VL-72B~\cite{qwen2-5-vl}, InternVL3.5-8B~\cite{internvl3}, and Qwen2.5-VL-7B~\cite{qwen2-5-vl}; (2) \textit{Medical Domain}: Lingshu-32B~\cite{lingshu}, MedVLM-R1~\cite{pan2025medvlm}, HuatuoGPT-V-7B~\cite{chen2024towards}, and Lingshu-7B~\cite{lingshu}; and (3) \textit{Thinking with Images}: GRIT~\cite{fan2026grit}. Lingshu-7B serves as both our foundation model and primary baseline. Training is conducted on 4$\times$A100 80GB GPUs using verl~\cite{sheng2025hybridflow}: SFT for 8 epochs ($lr=3e-6$, batch size 16), followed by GRPO for 2 epochs ($lr=1.5e-6$, $\beta=0.001$, $G$=6). Training completes within 24 hours.

\noindent\textbf{Evaluation Protocol.}
For localization and description, we adopt NOVA's evaluation metrics~\cite{bercea2026nova}. For diagnosis, we design an LLM-as-Judge framework where GPT-4o~\cite{gpt4o} evaluates three dimensions: \textit{Diagnostic Accuracy} (semantic match with ground truth), \textit{Reasoning Quality} (logical chain from imaging to diagnosis), and \textit{Safety} (clinical appropriateness). All scores are normalized to 0--100\%. Notably, our benchmark prompts provide only images and questions, whereas NOVA additionally incorporates clinical history.

\subsection{Results}


\begin{table}[t]
\centering
\caption{Anomaly Localization Results. We evaluate with detection metrics (mAP at IoU 0.30/0.50, reported as percentages), number of true positives (TP$_{30}$), and number of false positives (FP$_{30}$). Bold numbers indicate the best results among models of comparable scale ($\leq$30B).}
\label{tab:task1}
\setlength{\tabcolsep}{3pt}
\resizebox{\columnwidth}{!}{%
\begin{tabular}{lcccccccc}
\toprule
\multirow{2}{*}{\textbf{Model}} & \multicolumn{4}{c}{\textbf{Our Benchmark}} & \multicolumn{4}{c}{\textbf{NOVA (OOD)}} \\
\cmidrule(lr){2-5} \cmidrule(lr){6-9}
 & mAP$_{30}$ & mAP$_{50}$ & TP$_{30}$ & FP$_{30}\downarrow$ & mAP$_{30}$ & mAP$_{50}$ & TP$_{30}$ & FP$_{30}\downarrow$ \\
\midrule
\multicolumn{9}{c}{\textbf{\textit{General Domain}}} \\
\midrule
Qwen2.5-VL-72B & 55.16 & 25.56 & 1323/2384 & 624 & 37.02 & 21.25 & 397/1068 & 525 \\
LLaVA-v1.5-13B & 9.31 & 3.36 & 222/2384 & \textbf{140} & 9.55 & 4.12 & 102/1068 & \textbf{140} \\
InternVL3.5-8B & 21.18 & 10.63 & 645/2384 & 2106 & 11.40 & 3.75 & 125/1068 & 919 \\
Qwen2.5-VL-7B & 11.50 & 3.72 & 274/2384 & 1157 & 21.23 & 9.77 & 227/1068 & 651 \\
\midrule
\multicolumn{9}{c}{\textbf{\textit{Medical Domain}}} \\
\midrule
Lingshu-32B & 4.19 & 1.24 & 155/2384 & 2285 & 6.56 & 1.77 & 96/1068 & 1122 \\
HuatuoGPT-V-7B & 14.14 & 3.89 & 337/2384 & 647 & 15.07 & 4.49 & 161/1068 & 617 \\
MedVLM-R1 & 4.43 & 0.57 & 108/2384 & 1946 & 5.28 & 0.88 & 58/1068 & 831 \\
Lingshu-7B & 4.96 & 0.74 & 118/2384 & 1782 & 5.24 & 0.84 & 56/1068 & 836 \\
\midrule
\multicolumn{9}{c}{\textbf{\textit{Thinking with Images}}} \\
\midrule
GRIT & 15.73 & 6.27 & 353/2384 & 935 & 14.18 & 2.27 & 168/1068 & 742 \\
\midrule
\textbf{BrReMark (Ours)} & \textbf{64.93} & \textbf{37.54} & \textbf{1548/2384} & 812 & \textbf{33.05} & \textbf{13.30} & \textbf{353/1068} & 285 \\
BrReMark w/o synthesis & 60.32 & 35.00 & 1438/2384 & 911 & 29.87 & 8.82 & 319/1068 & 388 \\
BrReMark w/o $r_\text{loc}$ & 51.01 & 21.98 & 1216/2384 & 1161 & 23.25 & 6.93 & 251/1068 & 513 \\
BrReMark w/o $r_\text{llm}$ & 62.78 & 36.58 & 1509/2384 & 852 & 30.81 & 11.52 & 329/1068 & 310 \\
BrReMark w/o $r_\text{fmt}$ & 58.93 & 33.10 & 1405/2384 & 969 & 29.03 & 9.08 & 310/1068 & 459 \\
\bottomrule
\end{tabular}%
}
\end{table}

\noindent\textbf{Superior Open-Set (ID) Performance.}
Despite being compact (7B), BrReMark excels across localization, description, and diagnosis in ID scenarios (Table~\ref{tab:task1} and~\ref{tab:task23}). In anomaly localization, it achieves 64.93 mAP$_{30}$, outperforming 72B Qwen2.5-VL by 9.77. Precise grounding validates two-turn paradigm and provides explicit ROI for clinicians. For image description, BrReMark reaches a 21.57 Clinical F1, surpassing healthcare-adapted HuatuoGPT-V-7B (16.00) and trailing only Lingshu-32B. In clinical diagnosis, it achieves a leading Reasoning Quality score of 60.30. Powered by two-turn dataset, enhanced reasoning yields a state-of-the-art diagnostic accuracy of 45.26 (9.83 better than Lingshu-7B).

\noindent\textbf{Hallucination Suppression.}
Compared to Lingshu-7B, BrReMark achieves this through: 1) Hallucination Penalty reduces detection false positives (FP) from 1782 to 812; 2) Modality Gating improves description Modality F1 from 42.72\% to 47.68\%. They benefit downstream diagnosis, where safety rises from 61.64\% to 66.27\%, with accuracy 45.26\%, surpassing larger Lingshu-32B (45.07\%). 

\noindent\textbf{Generalization under Distribution Shift.}
BrReMark demonstrates exceptional OOD generalization, primarily driven by the use of synthetic anomaly data during reinforcement learning. In anomaly detection, our 7B model dramatically increases mAP$_{30}$ to 33.05 compared to Lingshu-32B (6.56), trailing only 72B Qwen2.5-VL due to parameter gap. For image description, it exhibits advantages against similar-sized models, improving the Clinical F1 (Clin) score from 13.84 (HuatuoGPT-V-7B) to 17.02. Although our synthetic data focuses on spatial localization rather than diagnostic semantics, it still yields improvements over the base Lingshu-7B, with the Safety score rising from 41.76 to 43.07. 

\begin{table}[t]
\centering
\caption{Image Description and Differential Diagnosis on Our Benchmark and NOVA. Description quality is evaluated by METEOR, Clinical F1, and Modality F1. Diagnosis quality is evaluated by Diagnostic Accuracy, Reasoning Quality, and Safety. All metric values are reported as percentages.}
\label{tab:task23}
\setlength{\tabcolsep}{2.5pt}
\resizebox{\columnwidth}{!}{%
\begin{tabular}{c|ccc|ccc|ccc|ccc}
\toprule
\multirow{3}{*}{\textbf{Model}} & \multicolumn{6}{c|}{\textbf{Description}} & \multicolumn{6}{c}{\textbf{Diagnosis}} \\
\cmidrule(lr){2-7} \cmidrule(lr){8-13}
& \multicolumn{3}{c|}{Our Benchmark} & \multicolumn{3}{c|}{NOVA (OOD)} & \multicolumn{3}{c|}{Our Benchmark} & \multicolumn{3}{c}{NOVA (OOD)} \\
\cmidrule(lr){2-4} \cmidrule(lr){5-7} \cmidrule(lr){8-10} \cmidrule(lr){11-13}
& MET & Clin & Mod & MET & Clin & Mod & Diag & Reas & Safe & Diag & Reas & Safe \\
\midrule
\multicolumn{13}{c}{\textit{\textbf{General Domain}}} \\
\midrule
Qwen2.5-VL-72B & 19.30 & 19.84 & 44.74 & 16.13 & 15.86 & 41.70 & 10.71 & 53.83 & 46.24 & 20.36 & 42.33 & 61.09 \\
LLaVA-v1.5-13B & 21.88 & 14.84 & 20.69 & 14.30 & 8.40 & 21.47 & 35.64 & 35.05 & 61.80 & 6.72 & 10.68 & \textbf{48.30} \\
InternVL3.5-8B & 20.52 & 14.86 & 30.95 & 17.18 & 15.51 & 57.49 & 9.73 & 27.34 & 28.89 & 11.20 & 26.04 & 40.85 \\
Qwen2.5-VL-7B & 18.67 & 18.18 & 41.83 & 14.50 & 14.32 & 37.84 & 7.15 & 42.23 & 43.15 & 6.28 & \textbf{29.35} & 30.23 \\
\midrule
\multicolumn{13}{c}{\textit{\textbf{Medical Domain}}} \\
\midrule
Lingshu-32B & 22.21 & 23.57 & 46.33 & 18.43 & 19.18 & 56.34 & 45.07 & 48.30 & 56.85 & 14.68 & 28.70 & 48.84 \\
HuatuoGPT-V-7B & 19.55 & 16.00 & 22.99 & 17.99 & 13.84 & 52.63 & 5.45 & 43.63 & 63.90 & 6.68 & 16.17 & 36.26 \\
MedVLM-R1 & 19.51 & 12.63 & 18.74 & 15.69 & 13.19 & 30.51 & 10.95 & 28.26 & 30.61 & 10.14 & 15.09 & 23.02 \\
Lingshu-7B & 17.06 & 21.47 & 42.72 & 17.26 & 15.84 & 54.36 & 35.43 & 48.25 & 61.64 & 11.09 & 21.03 & 41.76 \\
\midrule
\multicolumn{13}{c}{\textit{\textbf{Thinking with Images}}} \\
\midrule
GRIT & 19.66 & 20.80 & 25.69 & 15.74 & 13.34 & 35.47 & 20.20 & 26.62 & 33.30 & 6.80 & 12.52 & 32.23 \\
\midrule
\textbf{BrReMark (Ours)} & \textbf{23.64} & 21.57 & \textbf{47.68} & \textbf{18.04} & \textbf{17.02} & \textbf{61.30} & \textbf{45.26} & \textbf{60.30} & \textbf{66.27} & 10.43 & 26.32 & 43.07 \\
BrReMark w/o synthesis & 23.16 & \textbf{23.44} & 42.46 & 17.61 & 16.18 & 57.90 & 44.70 & 59.58 & 64.21 & \textbf{11.53} & 25.19 & 43.71 \\
BrReMark w/o $r_{\text{loc}}$ & 23.38 & 23.04 & 45.65 & 17.65 & 15.04 & 58.41 & 44.33 & 58.03 & 63.18 & 8.83 & 24.29 & 34.05 \\
BrReMark w/o $r_{\text{llm}}$ & 22.31 & 21.60 & 45.69 & 17.31 & 14.89 & 59.29 & 40.62 & 59.68 & 59.53 & 7.95 & 24.78 & 33.39 \\
BrReMark w/o $r_{\text{fmt}}$ & 23.57 & 21.48 & 42.91 & 17.44 & 15.27 & 59.22 & 44.35 & 59.62 & 63.16 & 8.90 & 24.11 & 35.65 \\
\bottomrule
\end{tabular}%
}
\end{table}

\noindent\textbf{Ablation Study.}
We conduct ablation studies (Tables~\ref{tab:task1} and \ref{tab:task23}) to validate key components. \textbf{(1)} The localization reward ($r_{loc}$) is crucial for spatial grounding; removing it causes localization performance on our benchmark to drop precipitously from 64.93 to 51.01. \textbf{(2)} Synthetic anomaly data is pivotal for OOD robustness, as excluding it degrades OOD localization from 33.05 to 29.87. \textbf{(3)} The semantic correctness reward ($r_{llm}$) is indispensable for text-heavy description and diagnosis tasks; omitting it significantly deteriorates diagnostic accuracy from 45.26 to 40.62. \textbf{(4)} The reasoning format reward ($r_{fmt}$) is vital for guiding the model to mimic human-like reasoning processes; discarding it diminishes overall performance, reducing diagnostic accuracy from 45.26 to 44.35.



\section{Conclusion}

We present BrReMark, the first grounded reasoning framework for open-ended brain MRI diagnosis. Through a hypothesis-mark-verification cognitive chain, BrReMark unifies anomaly detection, description, and diagnosis within a single interactive paradigm. Our multi-component reward function enables RL training on open-ended outputs, achieving substantial improvements on BrReMark-Bench (mAP$_{50}$: 0.74\%$\rightarrow$37.54\%) and strong generalization on NOVA.
Future work would consider to direct 3D volumetric reasoning and multi-modal integration (e.g., CT, other MRI sequences), with prospective clinical validation.

\subsubsection{Disclosure of Interests.}
The authors have no competing interests to declare that are relevant to the content of this article.

\bibliographystyle{splncs04}
\bibliography{Paper-3483}

\end{document}